\documentclass[journal]{IEEEtran}

\usepackage{graphicx}

\usepackage{amsmath}
\usepackage{amsfonts} 
\usepackage{subcaption}
\usepackage[style=ieee, backend=bibtex]{biblatex}
\usepackage[hidelinks]{hyperref}

\begin{document}
\title{A V2X-based Privacy Preserving Federated Measuring and Learning System}

\author{Levente~Alekszejenkó,
        Tadeusz~P.~Dobrowiecki%
\thanks{Department of Measurement and Information Systems,
Faculty of Electrical Engineering and Informatics,
Budapest University of Technology and Economics, 
Műegyetem rkp. 3., H-1111 Budapest, Hungary  e-mail: alelevente@mit.bme.hu, dobrowiecki@mit.bme.hu}%
}

\maketitle

\begin{abstract}
Future autonomous vehicles (AVs) will use a variety of sensors that generate a vast amount of data. Naturally, this data not only serves self-driving algorithms; but can also assist other vehicles or the infrastructure in real-time decision-making. Consequently, vehicles shall exchange their measurement data over Vehicle-to-Everything (V2X) technologies.
Moreover, predicting the state of the road network might be beneficial too. With such a prediction, we might mitigate road congestion, balance parking lot usage, or optimize the traffic flow. That would decrease transportation costs as well as reduce its environmental impact.

In this paper, we propose a federated measurement and learning system that provides real-time data to fellow vehicles over Vehicle-to-Vehicle (V2V) communication while also operating a federated learning (FL) scheme over the Vehicle-to-Network (V2N) link to create a predictive model of the transportation network.
As we are yet to have real-world AV data, we model it with a non-IID (independent and identically distributed) dataset to evaluate the capabilities of the proposed system in terms of performance and privacy. Results indicate that the proposed FL scheme improves learning performance and prevents eavesdropping at the aggregator server side.
\end{abstract}


\IEEEpeerreviewmaketitle

\section{Introduction}
While navigating through cities, autonomous vehicles (AVs) will be required to collect and process enormous amounts of data. This data contains raw measurement values (e.g., following distance) and derived observations (e.g., number of empty parking spaces on the spot). As these kinds of data further away from the trajectory of a vehicle might be hard to obtain, it is natural to exchange such knowledge among vehicles. The Vehicle-to-Vehicle (V2V) communication techniques can provide a channel for this information change. The received information might help the AVs in real-time operational (e.g., finding an empty parking lot) or tactical decision-making (e.g., heading towards an area that is supposed to offer many free parking spaces).

As the collected data, especially the derived ones, describes the actual traffic situation, AVs or even road operators shall access and process it. This data might support strategic planning that mitigates traffic congestion and optimizes the traffic flow (e.g., by building new parking lots or adjusting their pricing). Consequently, we might minimize the environmental impact of transportation. However, exchanging such enormous datasets over Vehicle-to-Everything (V2X) networks is challenging. We assume that this knowledge will be represented and distributed among the stakeholders (e.g., among AVs and road operators) as a machine learning model, e.g., the weight matrix of a neural network. Such representation has numerous benefits: it is a compressed extract of the data collected by the AVs, has predictive power, and can adapt to the changes in the road network. On a Vehicle-to-Network (V2N) channel, AVs can download the actual model from a central server and use its predictions for tactical planning (e.g., inferring the current parking lot occupancy rate in a neighborhood). AVs can also contribute to the training process of this model by participating in a federated learning (FL) system.

Unfortunately, AVs only have data about a relatively small area as they can only sense phenomena (e.g., parking lot occupancy) along their routes. Such data necessarily forms a not independent or identically distributed (non-IID) dataset. Non-IID data can cause performance issues and open the way for privacy threats for the FL system. In this paper, our essential contribution is \emph{addressing the problem of non-IID datasets by incorporating raw data sharing into the traditional FL scheme} for intelligent vehicles. As a proof of concept, we model the data of AVs by partitioning a well-known benchmark dataset MNIST \cite{mnist}. We assume raw data sharing is through a trusted (V2V) channel, while the participants access the FL server through a separate, untrusted (V2N) line. Based on this assumption, we demonstrate how raw data sharing improves learning speed and mitigates privacy issues.

\subsection{Related Works}
It will be vital to use V2X communication to exploit the merits of AVs. Connected vehicles can implement platoons and minimize the following distance \cite{swaroop, boubarki}, or make it possible to improve intersection management \cite{namazi}. As the V2X channel bandwidth is limited, we shall consider how to reduce the number of exchanged data packages. According to \cite{takamasa}, channel usage can be optimized by estimating and maximizing the information gains over the observations. This approach is theoretically related to the one described in this paper, but here, we aim to balance the datasets of the AVs to mitigate the effects of non-IID data.

Despite the numerous security mechanisms of V2X communication \cite{yoshizawa}, handling non-IID data on the application level is still an important issue. The non-IID nature of the data might disclose information about a vehicle's route revealing the origins and destinations of the passengers. In the case of commercial vehicles, that would mean that the whole supply chain of a company might be exposed. To mitigate this problem, there are both low-level solutions, such as using pseudonyms \cite{alani, yoshizawa} and certificates \cite{yoshizawa}, as well as application-level solutions, e.g., maximizing the location entropy in social internet of vehicle applications \cite{xing}.

The abovementioned solutions require a trusted third (or even fourth) party. Federated learning~\cite{original_fl}, that has gained popularity recently, may provide a simpler solution by an adoption to the vehicular domain. However, FL needs to exchange data periodically, which might be challenging in V2X communication; \cite{song} proved that FL works even in poor networking conditions. Although multiple papers \cite{song, wang} refer to FL as a privacy-protecting scheme, the survey in \cite{yin:survey} demonstrates how difficult it might be to secure an FL system. Most of the privacy-preserving FL mechanisms build on either distorting the data or classical cryptographic approaches that might be vulnerable to quantum computer algorithms. Considering the long lifespan of a vehicle or a piece of infrastructural equipment, we shall consider the post-quantum cryptography problem, which might be impossible with traditional solutions according to the survey of \cite{yoshizawa}. FL has vulnerabilities against either active or passive attacks in the V2X domain. While \cite{yamany} mitigates the impact of an active attack against the FL system; in this study, as an elaboration of our concept \cite{minisymp, mnist_old}, we focus on a passive eavesdropping attack against an FL scheme at a V2N server side and its prevention by trusted V2V communication.

The rest of the paper is organized as follows: In Section~\ref{sec:architecture}, we introduce the fundamental idea of a V2X-based, privacy-aware FL scheme. After that, Section~\ref{sec:eval} presents the evaluation of the proposed learning scheme. In Section~\ref{sec:data}, we describe how partitioning the MNIST data models obtained data of commuting AVs. Our key contribution is Section~\ref{sec:sharing}, which focuses on training data sharing among AVs. For numerical evaluation, Section~\ref{sec:meas} defines different measurement setups and Section~\ref{sec:train} elaborate the technical details of the training process. The defined measurement setups are compared by their learning performance in Section~\ref{sec:perf}, and by their privacy-awareness in Section~\ref{sec:threats}. Lastly, Section~\ref{sec:conclusion} concludes this paper.

\section{System Architecture}
\label{sec:architecture}
The principal assumption for our work is an extensive V2V information exchange between the vehicles. AVs can use the received data in real-time operational decision-making. Moreover, the retrieved knowledge (along with an AV's own measurements) can serve as training data for machine learning mechanisms. As V2V implementations are usually direct connections between two vehicles, we assume this channel is trusted and secure.

\begin{figure}[!t]
    \centering
    \includegraphics[width=.45\textwidth]{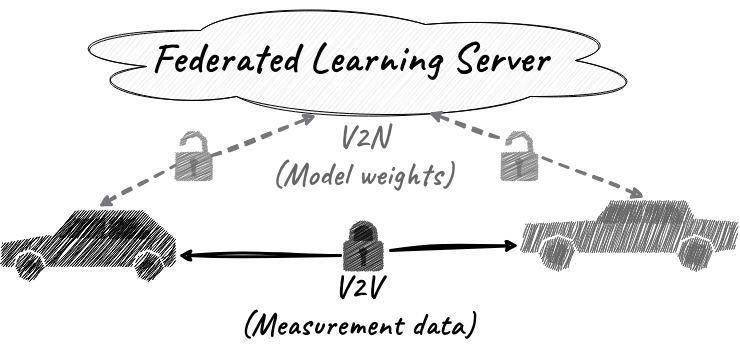}
    \caption{Illustration of the FL setup. Participant AVs can communicate with each other and the server. Participants can share data on a trusted V2V link, while the server receives and sends the model parameters of the FL process on an untrusted V2N channel.}
    \label{fig:setup}
\end{figure}

On the other hand, the community of AVs might measure some phenomena, e.g., congestion levels or occupancy rate of parking lots, that are beneficial to improve the traffic flow. Intelligent traffic controller systems, or even AVs can exploit the merits of such knowledge encapsulated in a queryable predictive model. To build such a model, we propose a vehicular FL system. The aggregator server of this system shall reside in the cloud, see Fig.~\ref{fig:setup}, and the \emph{participant} AVs shall connect to this server via V2N solutions and download the actual federated model. Additionally, the server can request a subset of the available AVs to train their local model and upload their updated weight matrix. The server runs a FedAvg \cite{original_fl} algorithm to create the next version of the federated model.

As the link between the AVs and the cloud is indirect, including, e.g., routers and switches, we might not safely assume that this link is secure. To prevent a successful eavesdropping attack against the vehicular FL system, we propose that AVs use their measurement data together with the information received over the V2V communication to moderate their non-IID characteristics. We expect this will improve the speed of the training process by potentially utilizing more data. Also, in this way, by reducing the non-IID nature of the training data of the participant AVs, it will lower the probability of a successful cyber attack. (It may also attenuate the bias of the optimal global model, but this problem is out of the scope of this paper.)

\section{Evaluation}
\label{sec:eval}
To evaluate the proposed FL system, we implemented a small model consisting of  10~participants (that represent the AVs) and an aggregation server. Whenever the server initiated a communication round, the participants exchanged a part of their training datasets (modeling the V2V communication). Then, each of the 10~participants ran exactly 1~epoch of training. Finally, the participants uploaded their new weight matrix to the server that aggregates the result by a FedAvg algorithm. Through this process, we can assess the learning process at a fine-grained level.

\subsection{Data Modeling}
\label{sec:data}

As it is a commonly used dataset in FL \cite{original_fl}, in adversarial machine learning literature \cite{qayyum}, and in the automotive or transportation domain \cite{song, yamany, chen, hammoud, kamp}, we used the MNIST \cite{mnist} dataset in the modeled training process. MNIST consists of hand-written digits (as $28 \times 28$ sized matrices) as input features, and the objective is to identify the corresponding digit as a 10-class classification problem.

\begin{figure*}[t!]
    \centering
    \begin{subfigure}[t]{0.3\textwidth}
        \centering
        \includegraphics[width=\textwidth]{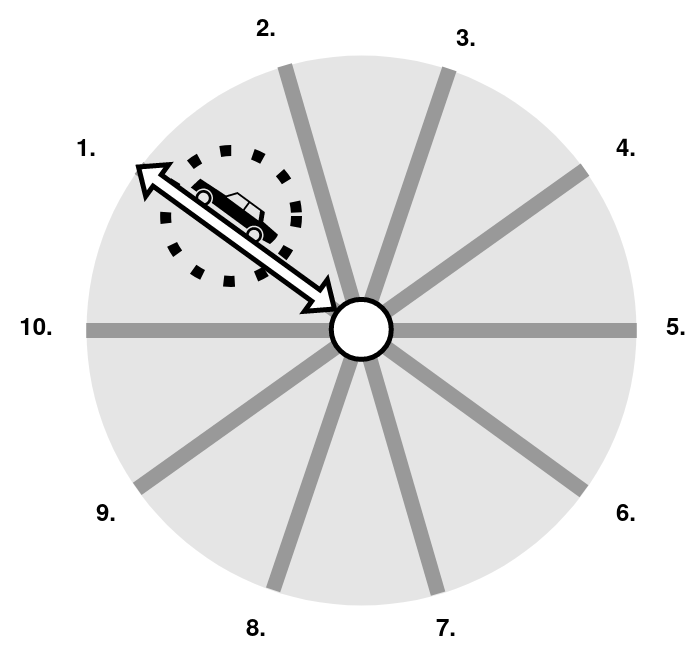}
        \caption{A vehicle commutes from the suburb to the city center along Avenue~1. During its movement, the vehicle measures a phenomenon in certain sensor range (dashed circle).}
        \label{fig:commute}
    \end{subfigure}\hfill
    \begin{subfigure}[t]{0.3\textwidth}
        \centering
        \includegraphics[width=\textwidth]{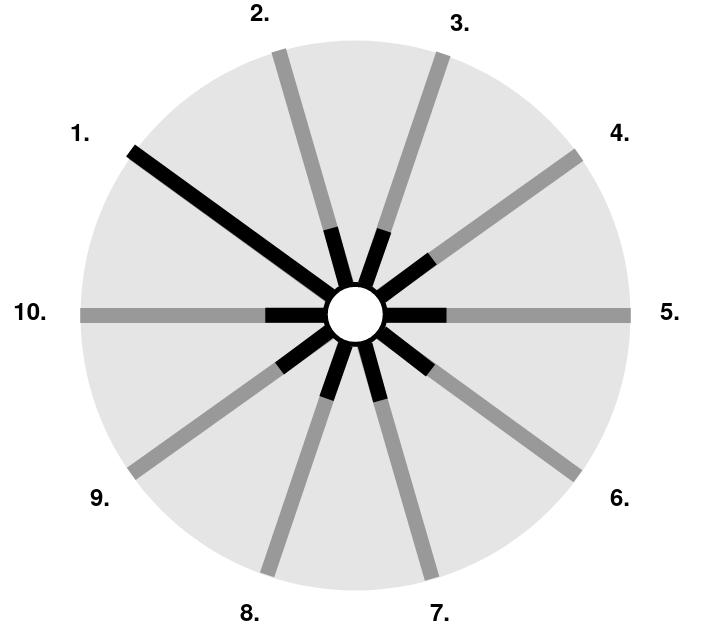}
        \caption{Consequently, the vehicle~1 will have many data from Avenue~1, and only a few from other avenues (from the black lines).}
        \label{fig:vehicledata}
    \end{subfigure} \hfill
    \begin{subfigure}[t]{0.3\textwidth}
        \centering
        \includegraphics[trim={0 0 0 1.25cm}, clip, width=\textwidth]{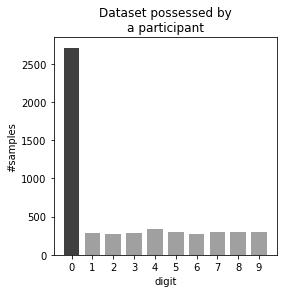}
        \caption{Dataset possessed by a participant. By partitioning the MNIST dataset, we can model the non-IID nature of the data of commuting vehicles.}
        \label{fig:mnist}
    \end{subfigure}
    \caption{Modeling commuting vehicles' data by partitioning the MNIST dataset.}
\end{figure*}

As non-IID data is the primary source of privacy issues, we partitioned the MNIST dataset into 10~parts. Each partition contained half of the samples belonging to a particular digit, and we distributed the remaining records of this digit randomly among the other 9~participants following a uniform distribution. This results in 10 non-IID datasets (of about the same size), and within each of these parts, there is a specific \emph{overrepresented} class label (a particular digit), see Fig.~\ref{fig:mnist}. The 10~modeled participants got these 10~non-IID datasets as training data (with approximately 5421~samples each\footnote{\label{note:mnist}The entire MNIST dataset contains approximately 60000~samples; however, digit~5 only has 5421~samples. Consequently, we used only 5421~samples from each digit label.}). Consequently, without balancing their datasets, every participant recognizes its overrepresented digit more accurately than the other numbers.

We consider this is an abstract model of a real-world scenario where 10~different cars commute from a suburb to the city center\footnote{The mentioned partitioning corresponds to a situation, e.g., when one commutes from the suburban town of Érd to the Inner City of Budapest, capital of Hungary. Then, in the city center, one moves in the 5--8\textsuperscript{th} Districts as well as in the 14\textsuperscript{th} District of Budapest.} along separate avenues and measure a phenomenon within a defined range, see Fig.~\ref{fig:commute} and measure data only in a specific part of the city (along their routes). As a result, the commuters will know their corresponding avenue well; therefore, measurements from their route form an overrepresented class in their dataset, see Fig.~\ref{fig:vehicledata}. Meanwhile, they will have only limited knowledge about the other parts of the city.

\subsection{Data Sharing}
\label{sec:sharing}
V2V data sharing is the prerequisite for operating the system described in Section~\ref{sec:architecture}. We assume that such communication technologies will have a limited range; therefore, in the modeled situation, vehicles can exchange data when they are close to each other, e.g., in the city center. We also wanted to model that AVs' recent measurements form possibly only a small dataset. Consequently, a small to-be-sent dataset was calculated for each vehicle in every communication round (simulated day, see Section~\ref{sec:train}). Then, the AVs transmitted this \emph{to-be-sent} dataset to each of the other 9~vehicles.

When the AVs get into communication range, they shall pick a piece of their raw data for sending. However, the exchanged raw data can also be an input for real-time decision-making; this aspect is out of the scope of the current paper. Hence, we assume that each record is equally important, and the vehicles can sample their data randomly upon knowledge exchange. This way, we aimed to balance each participant's dataset to be more IID and to test its effect on the proposed FL system.

To this end, we calculate how many samples from a digit category an AV shall receive from other vehicles in the beginning of each FL communication round. Let denote the number of digit categories as $n_c$, the number of participants as $n_p$, and the number of samples possessed by a participant from a given $i$ data class as $n_{s,i}$. Moreover, $p$ will be the rate of overrepresentation; i.e., the number of samples in the $o$ overrepresented class $n_{s,o}$ divided by the number of total records of the participant $\sum_{i=1}^{n_c} n_{s,i}$:

\begin{equation}
	p = \frac{n_{s,o}}{\sum_{i=1}^{n_c} n_{s,i}}.
 \label{eq:overrep}
\end{equation}

E.g., the data partitioning mentioned in Section~\ref{sec:data} results in $p=0.5$ overrepresentation rate. Using this notation, we can compute the $x$ number of samples belonging to each digit that an AV shall receive from all other participants to enhance the population of the underrepresented digit classes, and to make the AV's training dataset IID:

\begin{equation}
	\frac{n_s(1-p)}{n_c-1} + (n_p-1)x = \frac{n_s}{n_c}.
	\label{eq:data_balance}
\end{equation}

On the left-hand side of eq.~(\ref{eq:data_balance}), the first summand expresses the number of samples that a vehicle possesses (from an underrepresented class). The second term describes the number of samples that shall be received over V2V communication, and the right-hand side ensures the IID nature by a uniform distribution. Solving eq.~(\ref{eq:data_balance}) for an integer $x$ with the parameters of Table~\ref{tab:params}, yields $x = 27$. Consequently, each vehicle shall share $27$ samples from each digit category. As they have at least approx. $301$~records of each digit, they can choose $27$ out of it.

A higher $p$ indicates a less IID dataset (in our experiments, $p=0.1$ corresponds to the IID case, see Section~\ref{sec:meas}). Consequently, a higher $p$ would require more samples to share indicated by a higher $x$ value. At $p=1.0$, a participant would only possess samples from one data class, which would even make it impossible to run the data sharing method.

\begin{table}[!t]
\renewcommand{\arraystretch}{1.3}
\caption{Parameters to balance the AVs' data}
\label{tab:params}
\centering
      \begin{tabular}{|c|c|c|}
        \hline
        parameter & description & value\\ \hline
        $n_s$ & \#samples per category & $5421$ \\
        $n_c$ & \#categories & $10$ \\
        $p$ & overrepresentation rate & $0.5$ \\
        $n_p$ & \#participants & $10$ \\ \hline
      \end{tabular}
\end{table}

\subsection{Measurement Cases}
\label{sec:meas}
The original FL scheme performs well on non-IID parts built from the MNIST dataset \cite{original_fl}. Consequently, we suspect that raw data sharing will further improve the performance of the FL approach. In this section, we evaluate the learning performance of the proposed system.

To assess the merits of the proposed FL scheme enhanced by \emph{data exchange}, we will compare its capabilities to the following traditional approaches:
\begin{itemize}
	\item \emph{Centralized learning (baseline):} We trained a single convolutional neural network that uses the whole MNIST dataset for training.
	\item \emph{FL with IID data (IID):} We trained an original federated setup in which the participants had IID partitions of the MNIST dataset representing the idealistic data distribution of a federated system.
	\item \emph{FL with non-IID data (non-IID):} Without utilizing the V2V real-time data sharing, AVs are assumed to have non-IID data from a traffic phenomenon. To model this case, we trained a federated system with the non-IID ($p=0.5$) partitioning of the MNIST dataset described in Section~\ref{sec:data}.
\end{itemize}

\subsection{Training Process}
\label{sec:train}
To analyze how the training dataset distribution influences the model performance, the same $p=0.5$ MNIST partitioning was used for each FL measurement. We trained a simple convolutional neural network (CNN) described in \cite{example:mnist} within each experiment\footnote{We used the stochastic gradient descent (SGD) optimizer, and the hyperparameters were fixed: The learning rate was $0.01$, and the momentum was $0.9$.}. In the beginning, the weight matrix of the neural network was randomly initialized. To cope with the stochastic nature of neural network training, we repeated every measurement 10~times.

We used PyTorch to implement the CNN, while Flower \cite{flower} provided the framework for FL. Execution of the FL processes proceeds as follows:
\begin{enumerate}
	\item In case of the \emph{data exchange} measurements, each participant samples its to-be-sent $n_c \times x = 10 \times 27$ records (27 samples for each of the 10~digits) from its knowledge base. Then, the participants exchange among each other these to-be-sent datasets.
	\item To make a fine-grain evaluation possible, each participant trains its CNN for exactly 1~epoch. In the case of \emph{data exchange}, their training data is a mixture of their original partition and the received data; otherwise, they might only use their own MNIST portion.
	\item The participants send their updated model to the server of the FL system.
	\item The server aggregates the results with the FedAvg method.
	\item The server re-sends the updated model to the participants.
\end{enumerate}

In every measurement, we trained the CNN for 500~epochs.

The described training process models AVs departing from the suburbia in the morning and commuting to the city center. During their movements, they meet other AVs, with whom they share some information (gathered last evening and this morning) over a V2V communication solution. In the evening, the AVs return home; then, at night (e.g., while on charger), they train their CNN and send the model updates to the aggregating server over a V2N channel. Then, in the morning, the AVs can pull the latest federated model version from the service provider.

If a concrete use case required it, we would be able to increase the frequency of the FL rounds. A natural upper bound to this frequency is the frequency of meetings of AVs on the road.

\subsection{Performance Evaluation}
\label{sec:perf}
For evaluation, we will use the entire MNIST dataset (containing $m_i = 10 \times 5421$~samples\textsuperscript{\ref{note:mnist}}). This approach models a powerful attacker who is assumed to have access to all training data and whose goal is to identify what is the overrepresented class of an FL participant.

\begin{figure*}[!t]
    \centering
    \includegraphics[trim={0 0 0 .75cm},clip,width=.8\textwidth]{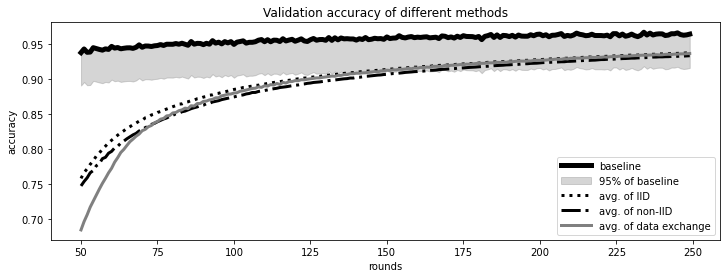}
    \caption{Test accuracy of the learning methods when reaching convergence, average of 10~measurements.}
    \label{fig:val_acc}
\end{figure*}

To evaluate the learning capabilities of the different measurement cases, we define validation \emph{accuracy} $a_i$ as rate of the $r_i$ number of successfully recognized samples over the number of used test samples $m_i$ after the $i$\textsuperscript{th} training run $a_i = \frac{r_i}{m_i}$. %

Now, we can measure the \emph{convergence speed} (CS) as a relative metric based on the accuracy. We compare the FL setup accuracy (the accuracy provided by the global, aggregated model) to the baseline case: CS is the first training round when the given method's accuracy falls within at least 95\% of the baseline accuracy. Thus, a lower CS value indicates that a learning process is faster.

Moreover, it is also crucial to check how accurate a FL system can be. To this end, we also measured the \emph{maximal accuracy} (MA) for each measurement case.  To mitigate the effect that the used CNN model might not fit the data perfectly, we publish MA values as the rate of the $\max_i a_i$ maximal accuracy over the maximal accuracy $\max_i a_{b,i}$ achieved by the baseline, centralized model as defined:

\begin{equation}
        \label{eq:ma}
    	\text{MA} = \frac{\max_i a_i}{\max_i a_{b, i}}.
\end{equation}

In Table~\ref{tab:perf}, we summarize the achieved performance parameters.

\begin{table}[!t]
\renewcommand{\arraystretch}{1.3}
\caption{Learning performance of the federated, global models; and security evaluation of model updates, success rate of an honest-but-curious attacker at the server side. Average of 10~measurements.}
\label{tab:perf}
    \centering
    \begin{tabular}{|l||cc|c|}
        \hline
        & \multicolumn{2}{c|}{\emph{learning performance}} & \emph{adversarial performance} \\
        \emph{method} & CS & MA & R\\ \hline
        \verb|IID| & $143$ & $98.77\%$ & 1.0 \\
        \verb|non-IID| & $163$ & $98.45\%$ & 10.0 \\
        \verb|data exchange| & $143$ & $98.65\%$ & 1.1\\ \hline
    \end{tabular}
\end{table}

According to Fig.~\ref{fig:val_acc} and Table~\ref{tab:perf}, data exchange improves the CS making the proposed system as fast as an ideal FL system, in which the participants possess IID distributed data portions, outperforming the non-IID system by approximately 14\% in CS. In the MA metric, the IID case is slightly better than the data exchanging and the non-IID approaches, but the differences are negligible.

\subsection{Privacy Threats}
\label{sec:threats}
Balancing the participants' datasets to be IID is beneficial for learning performance; however, the V2V data exchange might also function as a privacy-preserving technique for the FL.

Tracking a vehicle in the road network is unquestionably a privacy risk. Successful trackings may yield significant financial benefits or provide important intelligence reports. Considering the possible information gain, the technical challenges and the costs of such an attack, we can assume that the \emph{attacker} resides on the federated server side. Using the taxonomy of \cite{yin:survey}, we assume this attacker to be a passive, honest-but-curious attacker whose goal is to supply surveillance reports without disturbing the FL system. The attacker can eavesdrop on the V2N communication between the vehicles and the central federating learning server; hence, it accesses the reported model weight updates in the training phase.

For the evaluation, the attacker has access to the entire MNIST dataset. It is an unrealistic scenario, justified by the idea that if we can show that the proposed method preserves privacy even against such a strong attacker, it can also provide privacy in a more natural case.

The attacker aims to track a vehicle, i.e., in our model, to infer a participant's overrepresented data class. Hence, it presumes that the non-IID training dataset will result in a biased local model. Such a model predicts values from its overrepresented class more accurately, and its prediction accuracy is reduced in other parts of the feature space. Exploiting this assumption, the attacker evaluates the prediction accuracy of a participant's model described by the received weight updates as follows:

\begin{enumerate}
	\item The attacker samples $\mathcal{S}_0, \mathcal{S}_1, \dots \mathcal{S}_9$ evaluating records from the whole MNIST datasets for each corresponding digit.
	\item Given participant model $M$, the attacker calculates which is the most accurately predicted class on the average: $\hat{o} = \arg \max_i \bar{a}(M(\mathcal{S}_i))$, where $\bar{a}(M(\mathcal{S}_i))$ refers to the mean accuracy of the $M$ model on the $\mathcal{S}_i$ class.
	\item The attacker assumes that the participant had digit $\hat{o}$ as its overrepresented class.
\end{enumerate}

In our experiments, we had 10~participants (Participant\textsubscript{A}, Participant\textsubscript{B}, \dots, Participant\textsubscript{J}) that possessed digit~0, digit~1,~\dots,~digit~9 as $o_1, o_2,~\dots,~o_{10}$ overrepresented classes respectively. Let $R$ be the success rate of the attacker be measured as the number of correctly identified overrepresented classes $\sum_i I(o_i = \hat{o}_i)$, where $I(\cdot)$ stands for the indicator function being $1$ if its argument is true and $0$ otherwise. The attacker has a $p_s = \mathbb{P}(o = \hat{o}) = \frac{1}{n_c} = \frac{1}{10}$ chance to correctly identify the overrepresented class of a participant by random guessing in a single attack. Assuming the independence of the $n_a = 10$ attacks this process (following a binomial distribution) would result in an $R = n_a \cdot p_s = 1.0$ expected success rate.

\begin{figure}[!t]
    \centering
    \begin{subfigure}[t]{0.4\textwidth}
        \includegraphics[trim={0 0 0 1.25cm}, clip, width=\textwidth]{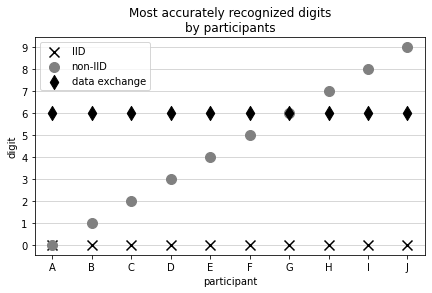}
        \caption{In 9 of the 10 training rounds the attacker cannot identify the overrepresented classes of the participants. An example of the usual performance of the attacker.}
    \end{subfigure}
    
    \vspace{6pt}
     
    \begin{subfigure}[t]{0.4\textwidth}
        \includegraphics[trim={0 0 0 1.25cm}, clip, width=\textwidth]{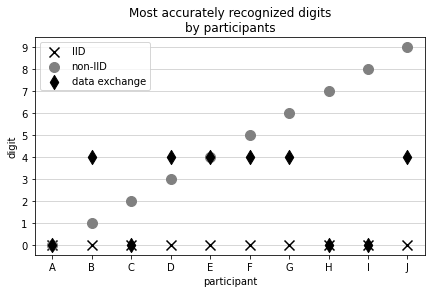}
        \caption{In 1 of the 10 training rounds the attacker might be more successful in identifying the overrepresented classes of the participants. An example of the best performance of the attacker.}
    \end{subfigure}
    
    \caption{Most accurately recognized overrepresented classes by the attacker. Results corresponds to single training processes with each method.}
    \label{fig:sec}
\end{figure}

Consequently, any $R \le 1.0$ indicates that the attacker is not more successful than random guessing, meaning that the participants (i.e., their weight updates) are algorithmically not distinguishable, even by an unnaturally strong attacker. As algorithmic indistinguishability is the general definition of security, we can consider an $R \le 1.0$ process to be secure.

Table~\ref{tab:perf} together with Fig.~\ref{fig:sec} summarizes the security evaluation results. The ideal IID case is secure by nature, and the attacker can only identify that participant\textsubscript{A} possesses the most samples of digit~0. This identification is only accidental (digit~0 turns to be the best estimate for each participant). The non-IID case, which corresponds to the traditional FL approach, is surprisingly revealing and has an $R = 10$ success. %
According to our experiments, incorporating measurement data from V2V communication into the training data sets of the participants helps to preserve users' privacy because the resulting process behaves almost identically (providing $R=1.1$ success without certainly distinguishing the participants) to the ideal IID one. Hence, such an attacker fails to firmly identify the overrepresented classes of the participants.

\section{Conclusion}
\label{sec:conclusion}
In this paper, we proposed a federated measuring and learning system architecture that utilizes the capabilities of V2X communication. Within this system, fellow vehicles share real-time information on V2V channels. In the meantime, the AVs train a federated machine learning model and communicate with a central aggregating server by V2N techniques. We propose that the AVs shall incorporate the real-time data exchanged on V2V into their training dataset to make it more independently and identically distributed. By adopting a benchmark machine learning problem, the handwritten digit recognition on the MNIST dataset, we provided experimental results that the \emph{proposed learning scheme improves the learning performance and, at the same time, mitigates the possible passive eavesdropping attacks against private information} in V2X communication systems. 

\subsection{Discussion}
To prove the security of the learning setup, we had to assume a powerful, honest-but-curious attacker on the FL site aiming to track the vehicles. In our experiment, this attacker had access to the entire training dataset, which is an unrealistic worst-case scenario. Moreover, the $p = 0.5$ MNIST partitioning provided articulated non-IID nature of the training data\footnote{Real-world or simulated vehicular datasets would add random noise to the distribution of records by nature: Traffic congestion modifies recording times. Depending on a concrete road network, realistic, traffic equilibria-based routing algorithms may provide various routes.}. We showed that despite such an advantage, in the majority of the cases, the attacker cannot distinguish the proposed solution (with raw-data exchanging participants) from an ideal one (with participants having IID data), which indicates that our approach can mitigate privacy threats against weaker and more realistic attackers.

Moreover, there are various FL aggregation algorithms that handle data heterogeneity (e.g., FedProx \cite{fedprox}, SCAFFOLD \cite{scaffold}, FedNova \cite{fednova}) and enhance model accuracy; but these methods do not guarantee privacy-preservation assuming the attacker can eavesdrop on the communication between the participants and the aggregator server. Hence, in this paper, we focused on alleviating the privacy threats at their origin by balancing the training datasets to have an IID nature.

Although, in Section~\ref{sec:sharing}, we sketched a simple data-sharing method; in real-world scenarios, to-be-sent data sets (with sample prioritization and handling of missing data) are harder to select. When calculating such a dataset, the sender might focus on achieving the highest possible data utility to provide information for the real-time decision algorithms of the receiver. However, in the meantime, the sender shall keep its private information hidden while minimizing also the usage of the communication channel. Solving these competing interests is a challenging task. In the future, our research will focus on solving this problem.

In this paper, the proposed learning system was illustrated by an application in the intelligent transportation domain. However, with proper adaption, it might be applied in different fields of study. To encourage experimentation and for further development, our source codes are publicly available on GitHub: \url{https://github.com/alelevente/FeLeSh}.

\section*{Acknowledgment}
Project no.~TKP2021-EGA-02 has been implemented with the support provided by the Ministry of Culture and Innovation of Hungary from the National Research, Development and Innovation Fund, financed under the TKP2021-EGA funding scheme. The project was also supported by the European Union project RRF-2.3.1-21-2022-00004 within the framework of the Artificial Intelligence National Laboratory. Supported by the ÚNKP-23-3-II-BME-233 New National Excellence Program of the Ministry for Culture and Innovation from the source of the National Research, Development and Innovation Fund.

\ifCLASSOPTIONcaptionsoff
  \newpage
\fi

\printbibliography

\end{document}